\documentclass{article}

\usepackage{arxiv}
\usepackage[utf8]{inputenc} 
\usepackage[T1]{fontenc}    
\usepackage{hyperref}       
\usepackage{url}            
\usepackage{booktabs}       
\usepackage{amsfonts}       
\usepackage{nicefrac}       
\usepackage{microtype}      
\usepackage{lipsum}		
\usepackage{graphicx}
\usepackage{natbib}
\usepackage{doi}
\usepackage{caption}
\usepackage{subcaption}
\usepackage{float}

\title{Adversarial Examples: Generation Proposal in the Context of Facial Recognition Systems}

\author{ 
        Marina Fuster  \\
	Department of Computer Science\\
	Instituto Tecnológico de Buenos Aires \\
        Buenos Aires, Argentina \\
	\texttt{mfuster@itba.edu.ar} \\
	\And
	Ignacio Vidaurreta  \\
	Department of Computer Science\\
	Instituto Tecnológico de Buenos Aires \\
        Buenos Aires, Argentina \\
	\texttt{ividaurreta@itba.edu.ar} \\
}

\date{October 11, 2021}

\renewcommand{\shorttitle}{Adversarial Examples: Generation Proposal in the Context of Facial Recognition Systems}
\hypersetup{
pdftitle={A template for the arxiv style},
pdfsubject={q-bio.NC, q-bio.QM},
pdfauthor={David S.~Hippocampus, Elias D.~Striatum},
pdfkeywords={First keyword, Second keyword, More},
}

\begin{document}
\maketitle

\begin{abstract}
In this paper we investigate the vulnerability that facial recognition systems present to adversarial examples by introducing a new methodology from the attacker perspective. The technique is based on the use of the autoencoder latent space, organized with principal component analysis. We intend to analyze the potential to craft adversarial examples suitable for both dodging and impersonation attacks, against state-of-the-art systems. Our initial hypothesis, which was not strongly favoured by the results, stated that it would be possible to separate between the "identity" and "facial expression" features to produce high-quality examples. Despite the findings not supporting it, the results sparked insights into adversarial examples generation and opened new research avenues in the area.
\end{abstract}

\section{Introduction}
The biometric systems field, particularly facial recognition technologies, has experienced significant advancement since the COVID-19 pandemic. An estimation by the research firm MarketsandMarkets, indicates that the global facial recognition market will grow from USD 3.8 billion in 2020 to USD 8.5 billion by 2025 (\cite{facial_recognition_market}).

Recent developments in facial recognition algorithms have shown remarkable improvements in accuracy. These algorithms have decreased their error rate from 4.1\% in 2014 to 0.08\% today. Not only the quality improvement has been incredible, but the number of algorithms achieving this precision has increased since then (\cite{facial_recognition_accuracy}).

Deep learning serves as the primary technology behind facial recognition systems. Despite their widespread use, deep neural networks exhibit a critical vulnerability: adversarial examples. A study published in 2014 identified these examples as inputs imperceptibly modified to cause misclassification when processed by a deep learning algorithm, such as a classification neural network. For instance, an adversarial example might lead a facial recognition system to misclassify "Brad Pitt" as "George Clooney" (\cite{intriguing_properties_of_neural_networks}).

Currently, there is no consensus on the underlying reasons for the existence of adversarial examples. In a study published in 2019, data characteristics are analyzed and categorized as either robust or non-robust (\cite{adversarial_examples_are_not_bugs}). Non-robust features are identified as a potential cause of adversarial examples, yet they are crucial for classifiers because removing these features results in a decline in accuracy.

Adversarial examples contribute evidence in favour of the transferability principle, as noted in \cite{intriguing_properties_of_neural_networks}: examples that disrupt one network's classification can similarly affect other networks, regardless of architectural differences or training datasets. This robustness against various classifiers enhances the importance of studying adversarial examples.

This paper presents a methodology for generating adversarial examples targeting facial recognition systems, utilizing autoencoders and principal component analysis. The analysis will explore not only the misclassification results but also the realism and quality of the potential examples.

\section{Related Work}

Two well-known methods for generating adversarial examples are the Carlini and Wagner (CW) method (\cite{robustness_neural_networks}) and the Fast Gradient Sign Method (FGSM) (\cite{harnessing_adversarial_examples}). These examples create adversarial examples by modifying an original element through perturbations. Additionally, Unrestricted Adversarial Examples are generated from scratch using generative algorithms, such as Generative Adversarial Networks (GANs) (\cite{unrestricted_adversarial_examples}). In this case, the imperceptibility of the modifications is less critical than the realism of the image.

The literature shows that adversarial examples are studied from both the attacker's and defender's perspectives. The attack methodologies are categorized into two modalities: white box, where the attacker has access to the network's architecture and weights (\cite{on_adveresarial_patches}), and black box, where the internal workings of the targeted network are unknown to the attacker (\cite{practical_black_box_attacks}).

Among the technologies employed, Principal Component Analysis is frequently used in both attack and defense strategies (\cite{principal_component_adversarial_example}), while the use of autoencoders is used predominantly in defense. For instance, the PuVAE (Purifying Variational Autoencoder) is employed to cleanse adversarial examples of their imperceptible modifications (\cite{puvae_purification}).

Studies on the construction of adversarial examples display a lack of formal quality analysis in the generated images. While some adversarial examples involve subtle modifications (\cite{harnessing_adversarial_examples}), others, such as those generated using PCAE, appear to reduce image quality (\cite{principal_component_adversarial_example}).

\section{Methods and Materials}
\subsection{Facial Recognition System}
Let $i : i \in I$ be an image and $I$ a set of digital faces. Let $t : t \in T$ be a label corresponding to the identity of a person. Then, $O : I \to T$ is a function that assigns the correct label.

Let $S : I \to R$ be a facial recognition system with a set $I_s$ of stored images against which comparison is performed. Then, $S(i) = r$, where $r$ is a vector of probabilities indicating $i$ similarity with stored images in $S$. 

\subsection{Attack}
In this study, an "attack" is defined as the process of submitting a modified input to a facial recognition system. Specifically, the term "attack" throughout the rest of this document will refer exclusively to the black box approach. As explained before, in a black box setting, the internal details of the facial recognition system are unknown.

\subsubsection{Dodging}
Dodging attacks occur when the average probability given by $S(i)$ for an image $i$ with a label $t$ falls below 80\% for those images $i_s \in I_s$ where $O(i_s)=t$.

\subsection{Impersonation}
Let $t$ be $O(i)$ for an image and let $t_i$ be any label different from $t$. Impersonation attacks occur when the average probability given by $S(i)$ is over 80\% for the set of images $i_s \in I_s$, where $O(i_s) = t_i$ verifies.

\subsection{Dataset}

\begin{figure}[H]
  \centering
  \includegraphics[width=0.5\textwidth]{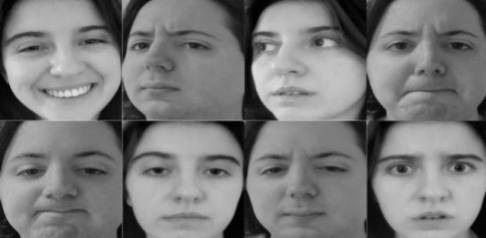}
  \captionsetup{width=0.5\textwidth}
  \caption{Sample images from dataset.}
  \label{fig:datasets}
\end{figure}

The dataset is manually built, reduced to a two-individual problem. One hundred photos were taken with cellphones, for each individual. Then, the MTCNN algorithm was used to detect the face and cut each image. All samples were scaled down to 256x256 and converted from color to black and white (single channel). A subset of the resulting photos can be found in \textit{Figure \ref{fig:datasets}}

The reasons behind these transformations were hardware limitations and the need to simplify the input to focus on the, if existing, identity feature.

\subsection{Proposed experimentation methodology}

\begin{figure}[H]
  \centering
  \includegraphics[width=0.8\textwidth]{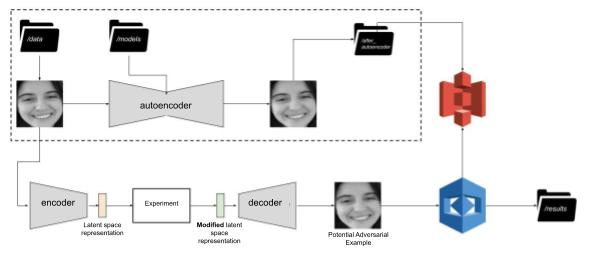}
  \captionsetup{width=0.8\textwidth}
  \caption{Diagram flow for experimentation mechanism.}
  \label{fig:experimentation_diagram}
\end{figure}

The diagram shown in \textit{Figure \ref{fig:experimentation_diagram}} has a two-step process. The first step, framed in a dotted line, will be executed only once: The images from the dataset are sent to the autoencoder to obtain their reconstructed counterparts. This set of images $I_s$ is then stored in the facial recognition system $S$: Amazon Rekognition. Since the autoencoder is not state-of-the art, it is a must to compare against reconstructed images to avoid similarity loss caused by quality loss and thus, misinterpreting the effect of modifications.

On the other hand, the second step which corresponds to the diagram outside the dotted line is rerun for every experiment. A sample from the dataset is chosen and the encoder is used to obtain the latent space representation of the image.

Principal component analysis is then performed on the vector obtained from the network latent representation. During the experiment phase, any necessary modifications are performed. Once finished, the inverse PCA operation is applied and the modified vector, which belongs again to the latent representation, is sent to the decoder for another feed-forward operation.

The resulting image is the potential adversarial example. It is then sent to Amazon Rekognition and the results of comparing that modified sample with the stored dataset $I_s$ are saved for further analysis.

After obtaining the results, post processing is made to understand if either dodging or impersonation attacks were successful. Since there is no access to the facial recognition system functionality, the attacks performed during the research are considered to be black box.

\subsubsection{Facial Recognition System Metrics}
Since the intention is to keep a focus not only on the similarity between subjects but also on the quality of the crafted adversarial examples, the following metrics are used from the comparison result. \textbf{Similarity}, which is the probability for the input image to match the stored image according to Amazon Rekognition. The \textbf{Confidence} metric is also a value between zero and one, indicating how confident Amazon Rekognition is about detecting a face in the image. Finally, \textbf{Quality Sharpness} and \textbf{Quality Brightness} will be used to calculate baselines to evaluate adversarial examples quality. It will be useful to understand if there was any quality loss during experimentation.

\section{Results}

\begin{figure}[H]
  \centering
  \includegraphics[width=0.9\textwidth]{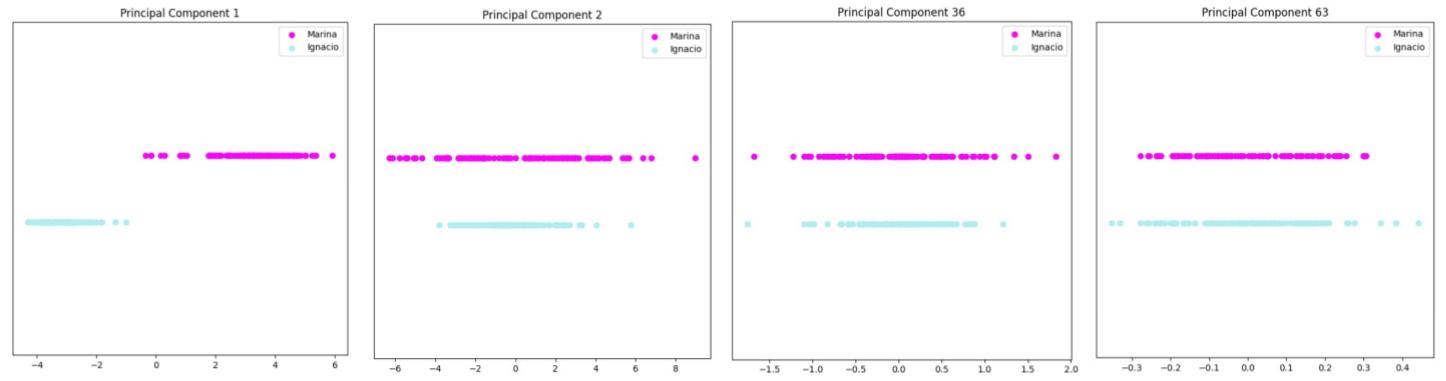}
  \captionsetup{width=0.9\textwidth}
  \caption{Four representative principal components out of sixty-four}
  \label{fig:pca_dimensions}
\end{figure}

Out of processing the entire dataset and posterior principal component analysis, the only component that displayed clear separation between subjects was the first component, as show in \textit{Figure \ref{fig:pca_dimensions}}. This result could be an indicative that modifying the value corresponding to PC1 for an image $i$ could modify. $O(i)$. On the other hand, modifying the other components could change other features, such as the expression.

In order to test this, the average image for all the images with label Ignacio and the ones with label Marina were taken and, by modifying the value of the first component, it was verified the progression towards the identity of their counterpart. Upon verifying this, more evidence was added to the fact that PC1 could control the label assigned by $O$, the identity.

\begin{figure}[H]
  \centering
  \includegraphics[width=0.8\textwidth]{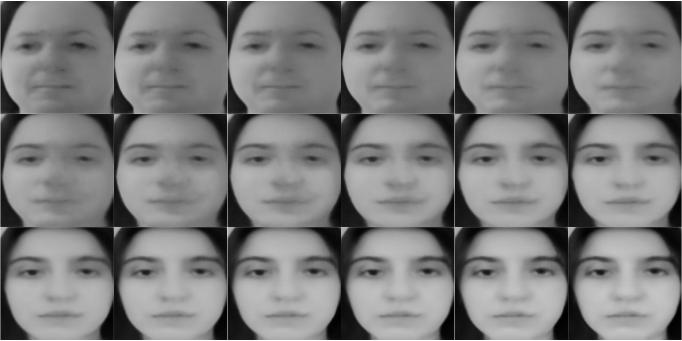}
  \captionsetup{width=0.8\textwidth}
  \caption{Transition from average "Ignacio" first principal component value towards "Marina" first component value.}
  \label{fig:transition_first_component}
\end{figure}

\begin{figure}[H]
  \centering
  \includegraphics[width=0.7\textwidth]{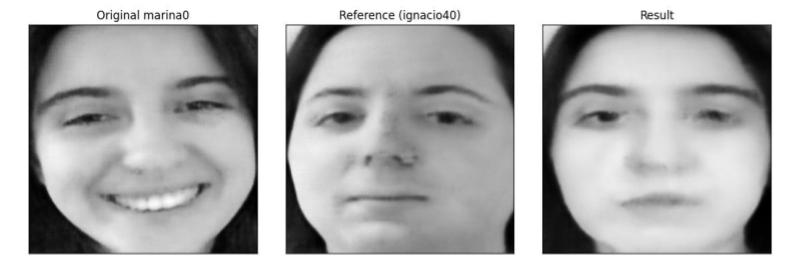}
  \captionsetup{width=0.6\textwidth}
  \caption{Modification where original principal components' representation gets modified with reference values for all expected the first component.}
  \label{fig:reference_and_original}
\end{figure}

On the other hand, to test the fact that the remaining sixty-three components controlled the features of the face, original and reference images were chosen to impose the reference's facial expression upon original's identity, which is shown in \textit{Figure \ref{fig:reference_and_original}}. Further analysis upon the semantic meaning on individual principal components showed no promise of a subset of values controlling things such as orientation, smile, eye-opening expression, etc.

In order to obtain clean results, baseline metrics were computed to understand the best possible values that Rekognition could return 
because different images, even for the same person, did not return 100\%, upon inspection. First, each image of Marina was compared against Rekognition's bank of images and the average and standard deviation for each result was obtained. It is important to note that the images used for the comparison are the same as the ones stored in the bank. Afterwards, the same process was applied with the images with Ignacio.

\begin{table}[H]
    \centering
    \begin{tabular}{ |p{3cm}||p{3cm}|p{3cm}|p{3cm}|p{3cm}|  }
     \hline
     \multicolumn{5}{|c|}{Baseline Values for Rekognition} \\
     \hline
     Target & Marina Mean & Marina Std Dev. & Ignacio Mean & Ignacio Std Dev. \\
     \hline
     Marina & 99,726 & 0,839 & 1,845 & 2,185 \\
     Ignacio & 1,843 & 2,004 & 98,945 & 3,015 \\
     \hline
    \end{tabular}
\end{table}

\begin{table}[H]
    \centering
    \begin{tabular}{ |p{3cm}||p{3cm}|p{3cm}|  }
     \hline
     \multicolumn{3}{|c|}{Baseline Values for Rekognition} \\
     \hline
     Metric & Mean & Std Dev. \\
     \hline
     Confidence & 99,999 & 1,48E-05 \\
     Quality Brightness & 83,025 & 5,978 \\
     Quality Sharpness & 40,989 & 8,596 \\
     \hline
    \end{tabular}
\end{table}

When evaluating quality metrics, images had to verify baseline values. All potential adversarial examples failing to corroborate this, were eliminated from the analysis.

\begin{figure}[H]
  \centering
  \includegraphics[width=0.7\textwidth]{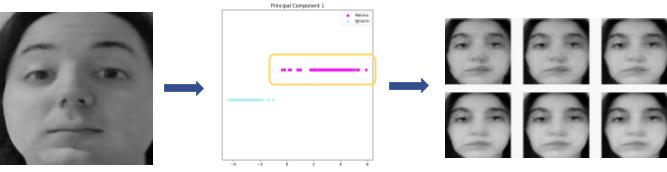}
  \captionsetup{width=0.6\textwidth}
  \caption{Potential Adversarial Examples Generation Technique.}
  \label{fig:edition_example}
\end{figure}

Afterwards, several experiments were conducted to craft potential adversarial examples. In order to do this, as shown in \textit{Figure \ref{fig:edition_example}}, several images from the reconstructed dataset were taken. There were two experiments' categories: those modifying the first principal component value (to change the identity) and those modifying the first three principal components' values (to change both identity and facial expression). In both types of experiment, the range in which modifications were going to happen were taken for each principal component separately. Then, a step was used to generate multiple images within those ranges.

Using this technique, it was possible to obtain dodging adversarial examples and impersonation adversarial examples. A survey of 51 individuals decided the label of the modified images and, with those results, we were able to defined to oracle's result.

\begin{figure}[H]
    \centering
    \begin{subfigure}[b]{0.4\textwidth}
        \includegraphics[width=\textwidth]{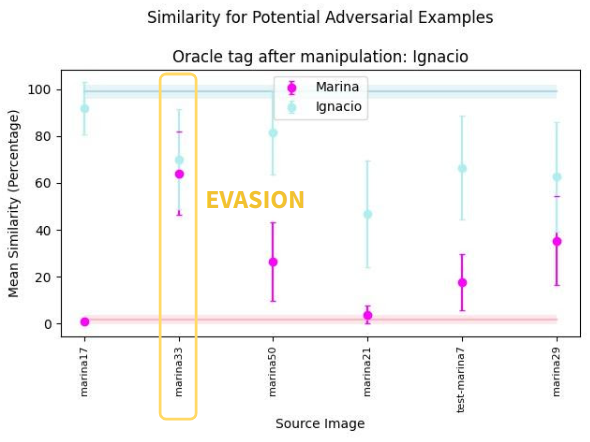}
        \caption{Dodging Attack}
    \end{subfigure}
    \begin{subfigure}[b]{0.4\textwidth}
        \includegraphics[width=\textwidth]{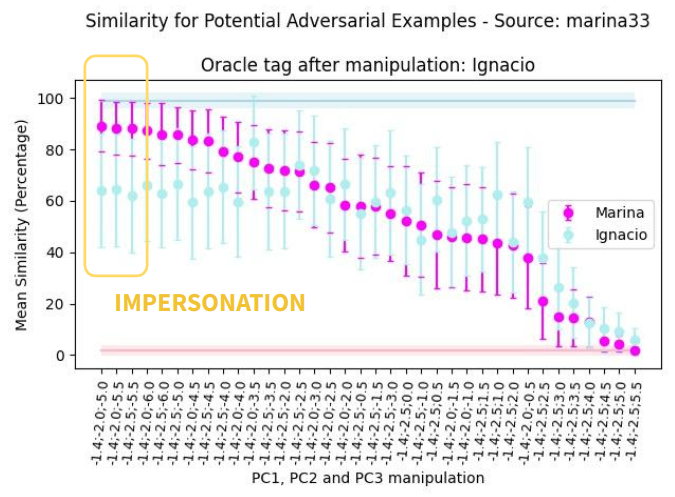}
        \caption{Impersonation Attack}
    \end{subfigure}
\end{figure}

On the other hand, a brute-force approach was also tested: we randomly selected two images and modified the first image according to every principal component value from the second image. Then, some images were manually selected and sent to Rekognition. Without paying attention to which principal component contributed which feature, it was possible to obtain adversarial examples here as well.

\section{Discussion}

This research highlights several critical issues surrounding the proposed approach. Firstly, the technique employed to obtain adversarial examples proved inconsistent, requiring extensive manual intervention. The inability to systematically identify why certain images were suitable as adversarial examples underscores the need for a more robust methodology.

Secondly, the quality of the adversarial examples produced in this study was highly questionable. Although these examples met established benchmarks, stricter criteria should have been applied to ensure higher fidelity. This discrepancy highlights a potential overemphasis on meeting benchmarks at the expense of practical applicability.

The restriction of the study to only two individuals may have also skewed the results. Notably, the expressive features in Ignacio's images, as opposed to Marina's more neutral expressions, suggest that expression features may be intricately linked to a person's identity. This observation indicates that a broader, more diverse dataset might offer different appreciations into the interaction between facial expressions and identity features.

Future research should therefore focus on expanding the dataset to determine whether the first principal component consistently acts as a "signal" for identity across a broader array of subjects. Moreover, investigating the underlying reasons for how relationships are modeled in this context could constitute a significant area of study, aimed at understanding the dynamics of adversarial example generation.

While the methodology discussed here offers a cost-effective approach for rapid adversarial attacks, its applicability is severely limited under specific conditions. This raises questions about the practical utility of the approach.

Lastly, the implementation of adversarial examples suggest that these techniques might only be relevant in particular scenarios, such as content manipulation to get around ad-blocking software. This perspective challenges the current level of attention adversarial examples are receiving within the academic and security communities (\cite{list_ae_papers}). The lack of consensus on the very reason of adversarial examples generation complicates the development of effective defense mechanisms. This ongoing debate highlights a fundamental gap in our understanding of neural networks threats and add emphasis to the need for continued research in this area.

\section{Conclusion}

Although this study did not identify a systematic method that enables a realistic adversarial attack scenario, the experimental methodology with an iterative approach—encompassing hypothesis formulation, procedural execution, and detailed analysis—proved invaluable for investigating adversarial examples and is recommended for future research in related areas. While the principal component analysis utilized to organize the latent space did not provide conclusive evidence supporting its effectiveness for the intended evasion and impersonation attacks, alternative techniques, such as Kohonen networks or variational autoencoders, may offer more promising results. Moreover, the study's limited applicability and the inconsistent quality of adversarial examples generates a call for a broader and more rigorous approach to better understand and enhance the robustness and applicability of adversarial techniques in practical scenarios.

\bibliographystyle{unsrtnat}
\bibliography{thesis}

\end{document}